\documentclass[10pt,a4,oneside,twocolumn]{article}

\usepackage{hyperref}
\usepackage{multicol}
\usepackage{setspace}
\usepackage[margin=2cm]{geometry}
\usepackage[utf8]{inputenc}
\usepackage[T1]{fontenc}
\usepackage[tracking]{microtype}

\usepackage{color,graphicx}
\usepackage[usenames,dvipsnames]{xcolor}

\usepackage{mathtools} 
\usepackage{amssymb}   
\usepackage{amsfonts}  
\usepackage{mathrsfs}  
\usepackage[group-separator={,}]{siunitx}   
\usepackage{numprint}  
\usepackage[capitalise,nameinlink]{cleveref} 
\usepackage{csquotes} 
\usepackage{textcase}
\usepackage[explicit,pagestyles]{titlesec}
\usepackage{parskip} 
\usepackage{xparse}
\usepackage{subcaption}
\usepackage[noend]{algpseudocode}
\usepackage[]{algorithm}

\usepackage[sc,osf]{mathpazo}   
\usepackage[euler-digits,small]{eulervm}

\usepackage[
	backend=bibtex,
	bibencoding=utf8,
	style=alphabetic,
	doi=false,isbn=false,url=false,
	firstinits=true,	
	dateabbrev=false,	
]{biblatex}
\let\citep\autocite
\let\cite\autocite

\usepackage[thmmarks]{ntheorem}
\crefname{Theorem}{Theorem}{Theorems}
\crefname{Corollary}{Corollary}{Corollaries}
\crefname{Lemma}{Lemma}{Lemmas}
\crefname{Definition}{Definition}{Definitions}
\crefname{Proposition}{Proposition}{Propositions}
\crefname{Problem}{Problem}{Problems}
\crefname{Proof}{Proof}{Proofs}
\crefname{Problem}{Problem}{Problem}

\makeatletter
\newtheoremstyle{msplain}%
{\item[\hskip\labelsep \theorem@headerfont ##1\ ##2\theorem@separator]}%
{\item[\hskip\labelsep \theorem@headerfont ##1\ ##2\ \normalfont{(\textit{##3})}\theorem@separator]}

\newtheoremstyle{msnonumberplain}%
{\item[\hskip\labelsep \theorem@headerfont ##1\theorem@separator]}%
{\item[\hskip\labelsep \theorem@headerfont ##3\theorem@separator]}%
\makeatother

\theoremnumbering{arabic}
\theoremstyle{msplain}
\theoremsymbol{\guillemotleft}
\theorembodyfont{\normalfont}
\theoremheaderfont{\normalfont\sffamily\bfseries}
\theoremseparator{:}
\theorempreskip{.5em} 
\theorempostskip{0em} 
\newtheorem{Theorem}{Theorem}
\newtheorem{Proposition}{Proposition}

\newtheorem{Corollary}{Corollary}

\newtheorem{Definition}{Definition}

\theoremstyle{msnonumberplain}
\theoremheaderfont{\normalfont\sffamily\bfseries}
\theorembodyfont{\normalfont}
\theoremsymbol{\ensuremath{_\Box}}
\newtheorem{Proof}{Proof}
\qedsymbol{\ensuremath{_\Box}}

\algrenewcommand{\alglinenumber}[1]{\color{black!50}\sf\footnotesize#1:}
\algrenewcommand\algorithmicrequire{\textbf{Require:}}
\algnewcommand\algorithmicinit{\textbf{Initialize:}}
\algnewcommand\Initialize{\item[\algorithmicinit]}
\algnewcommand\algorithmicout{\textbf{Result:}}
\algnewcommand\Output{\item[\algorithmicout]}
\algnewcommand\algorithmicmain{\textbf{Algorithm:}}
\algnewcommand\Main{\item[\algorithmicmain]}

\DeclareRobustCommand{\spacedallcaps}[1]{\sffamily{\MakeTextUppercase{#1}}}%
\DeclareRobustCommand{\spacedlowsmallcaps}[1]{\sffamily{\scshape\MakeTextLowercase{#1}}}%

\titleformat{\section}{\raggedright\normalfont\Large\sffamily}{\thesection}{.5em}{{#1}}
\titleformat{\subsection}{\raggedright\normalfont\sffamily}{\thesubsection}{1em}{\spacedallcaps{#1}}
\titleformat{\subsubsection}{\normalfont\sffamily}{\thesubsubsection}{1em}{\spacedallcaps{#1}}
\titleformat{\paragraph}{\normalfont\normalsize\sffamily}{\textsc{\MakeTextLowercase{\theparagraph}}}{0pt}{\spacedlowsmallcaps{#1}} 

\graphicspath{{./fig/}}

\definecolor{webgreen}           {rgb}{0   ,.5   ,0    }
\definecolor{webbrown}           {rgb}{.6  ,0    ,0    }
\hypersetup{%
    colorlinks=true, 
    urlcolor=webbrown, linkcolor=black, citecolor=webgreen
}

\DeclarePairedDelimiter\norm{\lVert}{\rVert}                 
\DeclarePairedDelimiter\ip{\langle}{\rangle}				 
\newcommand{\pmeasure}[1]{\ensuremath{\mathbb{#1}}} 
\newcommand{\pmeasureP}{\pmeasure{P}} 
\let\PM\pmeasureP
\newcommand{\Real}   {\ensuremath{\mathbb{R}}} 
\newcommand{\Nat}    {\ensuremath{\mathbb{N}}} 
\newcommand*{\diffdchar}{d}									 
\newcommand*{\dd}{\mathop{\diffdchar\!}}
\DeclareMathOperator*{\argmin}{arg\,min}
\DeclareMathOperator{\dom}{\ensuremath{dom}}
\newcommand{\im}{\imath} 
\providecommand\given{}
\DeclarePairedDelimiterX\ExpX[1]{[}{]}{
\renewcommand\given{  \nonscript\:
  \delimsize\vert
  \nonscript\:
  \mathopen{}
  \allowbreak}
#1
}
\newcommand\Exp{\mathbb{E}\ExpX}

\DeclarePairedDelimiterX\SetDefX[1]{\lbrace}{\rbrace}{
\renewcommand\given{  \nonscript\:
  \delimsize\vert
  \nonscript\:
  \mathopen{}
  \allowbreak}
#1
}
\newcommand\SetDef{\SetDefX}

\begin{document}

\title{\normalfont\spacedallcaps{Constant Time EXPected Similarity Estimation using Stochastic Optimization}}
\author{\spacedlowsmallcaps{Markus Schneider\textsuperscript{$\star\dagger$}, Wolfgang Ertel\textsuperscript{$\dagger$} \& G\"unther Palm\textsuperscript{$\star$}}}
\date{}

\twocolumn[
\begin{@twocolumnfalse}
\maketitle
\hrule
\section*{Abstract}
A new algorithm named \emph{EXPected Similarity Estimation} (EXPoSE) was recently proposed to solve the problem of \emph{large-scale anomaly detection}. It is a non-parametric and distribution free kernel method based on the Hilbert space embedding of probability measures. Given a dataset of $n$ samples, EXPoSE needs only $\mathcal{O}(n)$ (linear time) to build a model and  $\mathcal{O}(1)$ (constant time) to make a prediction. 
In this work we improve the \emph{linear} computational complexity and show that an $\epsilon$-accurate model can be estimated in \emph{constant} time, which has significant implications for large-scale learning problems.
To achieve this goal, we cast the original EXPoSE formulation into a stochastic optimization problem.
It is crucial that this approach allows us to determine the number of iteration based on a desired accuracy $\epsilon$, \emph{independent of the dataset size} $n$. We will show that the proposed stochastic gradient descent algorithm works in general (possible infinite-dimensional) Hilbert spaces, is easy to implement and requires no additional step-size parameters.
\vspace{1em}
\hrule
\end{@twocolumnfalse}
\vspace{2em}
]

{\let\thefootnote\relax\footnotetext{\raggedright\textsuperscript{$\star$} \textit{Institute of Neural Information Processing,\\
	University of Ulm, Germany}}}
{\let\thefootnote\relax\footnotetext{\raggedright\textsuperscript{$\dagger$} \textit{Institute for Artificial Intelligence,\\ 
    Ravensburg-Weingarten University of Applied Sciences, Germany}}}

 \section{Introduction}
 
 \emph{EXPected Similarity Estimation} (EXPoSE) was recently proposed to solve the problem of large-scale anomaly detection, where the number of training samples $n$ and the dimension of the data $d$ are too high for most other algorithms \autocite{Schneider2015a}. Here,
 \textcquote{Chandola2009}{anomaly detection refers to the problem of finding patterns in data that do not conform to expected behavior. These non-conforming patterns are often referred to as \emph{anomalies}}.
 
\begin{figure}[tb]
    \centering
    \includegraphics{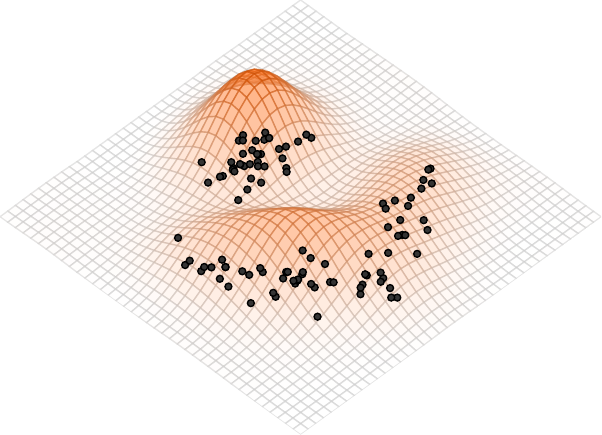}
    \caption{Sketch of the EXPoSE scores $\eta(y)$ in $\Real^2$, given some samples (black dots).}
    \label{fig:expose_synthetic_contour}
    \end{figure}

 As explained later in detail, the EXPoSE anomaly detection classifier
 \begin{align*}
 \eta(y) = \ip{ \phi(y), \mu[\PM]}
 \end{align*}
 calculates a score (the likelihood of $y$ belonging to the class of normal data) using the inner product between a feature map $\phi$ and the kernel mean map $\mu[\PM]$ of the distribution $\PM$ (\cref{fig:expose_synthetic_contour}). 
 Given a training dataset of size $n$, the authors provide a methodology to train this classifier in $\mathcal{O}(n)$ time and show that calculating a score for a query point can be done in $\mathcal{O}(1)$ time. The question arises if it is possible to improve on the linear training time and create an algorithm which is completely independent of the dataset size.
 
 The answer to this question is positive if a high accuracy sample estimate of $\mu[\PM]$ does not improve the anomaly detection performance. 
 As \citeauthor{bousquet2008tradeoffs}~\autocite{bousquet2008tradeoffs} observed, for most machine learning applications there is no need to optimize below the statistical error. The authors argue that accurately minimizing an empirical cost function does not gain much since it is itself an approximation of the expected costs and therefore contains errors. 
 We will see that it is possible to determine the number of samples needed to achieve a desired accuracy (the maximal deviation from the optimal model) of EXPoSE without any dependence on the datasets size~$n$.
 
 \subsection{Contributions \& Related Work}
 
 In this work we derive a methodology to build an $\epsilon$-accurate model $w$ of $\mu[\PM]$  using only a random subset of the training data by means of stochastic optimization.
  \begin{Definition}
 We say an algorithm finds an $\epsilon$-accurate solution $w$ of an objective function $f$ if
 \begin{align*}
 f(w) \leq \inf f + \epsilon
 \end{align*}
 for a given $\epsilon > 0$. 
 \end{Definition}
 We will show that for the proposed objective function $\Exp{f(w_t)-f(\mu[\PM])} \leq \mathcal{O}(1/t)$, where $w_t$ only needs access to $t$ random dataset elements, $t \in \SetDef{1,2,\dotsc,n}$. The key observation is that
 for a given an $\epsilon >0$ we can reach $\norm{w_t-\mu[\PM]} <\epsilon$ in a \emph{fixed} number of iterations \emph{independent} of the dataset size. Moreover, it can be shown that (without further assumptions) the $\mathcal{O}(1/t)$ rate is optimal for stochastic optimization \autocite{agarwal2009information}.
 
 Due to the low iteration costs, stochastic optimization and especially stochastic gradient (SG) methods \autocite{bousquet2008tradeoffs,roux2012stochastic}, are widely used for training machine learning models on very large-scale datasets. Such algorithms are used for example to train support vector machines \autocite{Shalev-Shwartz2007}, logistic regression \autocite{bach2014adaptivity} and lasso models \autocite{shalev2011stochastic}. However, \emph{this is the first time that EXPoSE is considered as an optimization problem} and we will show that the derived algorithms is of general interest for applications of the kernel mean map $\mu[\PM]$.
 
 Other optimization techniques such as projected gradient decent \autocite{boyd2004convex} or Nesterov's accelerated gradient descent \autocite{nesterov1983method,nesterov2004introductory} are also applicable in principle, however a single gradient evaluation takes already $\mathcal{O}(n)$ time and hence would be slower than the originally proposed EXPoSE approach. Other stochastic gradient methods \autocite{roux2012stochastic} can obtain a better convergence rate than $\mathcal{O}(1/t)$ for an objective composed of a sum of smooth functions. However this requires multiple passes over the datasets is therefor of no benefit.
 
 \section{Problem Description}
 EXPoSE is a probabilistic approach which assumes that the \emph{normal}, non-anomalous data is distributed according to some measure $\PM$. More formally, let $X$ be a random variable taking values in a measure space $(\mathcal{X},\mathscr{X})$ with distribution $\PM$. We denote the reproducing kernel Hilbert space (RKHS) associated with the kernel $k\colon \mathcal{X} \times \mathcal{X} \to \Real$ with $(\mathcal{H},\ip{\cdot,\cdot})$. A RKHS is a Hilbert space of functions $g\colon \mathcal{X} \to \Real$, where the evaluation functional $\delta_x\colon g \mapsto g(x)$ is continuous. The function $\phi\colon\mathcal{X} \to \mathcal{H}$ with
 \begin{align*}
 k(x,y) = \ip{\phi(x),\phi(y)} 
 \end{align*}
 is called \emph{feature map} denoted by $\phi(x) = k(x,\cdot)$.  Throughout the paper, we use $\norm{\cdot}_\mathcal{H}$ to denote the norm induced by the inner product defined as $\norm{g}_\mathcal{H} = \sqrt{\ip{g,g}}$.
 
 EXPoSE calculates a score which can be interpreted as the likelihood of a query point belonging to the distribution of normal data $\PM$. This is done in the following way.
 \begin{Definition}
 The \emph{expected similarity} of $y \in \mathcal{X}$ to the (probability) distribution $\PM$ is defined as
 \begin{align*}
 \eta(y) = \int_{\mathcal{X}} k(y,x) \,\mathrm{d}\PM(x), 
 \end{align*}
 where $k\colon\mathcal{X} \times \mathcal{X} \to \Real$ is a reproducing kernel.
 \end{Definition}
 Intuitively speaking the query point $y$ is compared to all other points of the distribution $\PM$. It can be shown \autocite{Schneider2015a} that this equation can be rewritten as an inner product between the feature map $\phi(y)$ and the kernel embedding $\mu[\PM]$ of $\PM$ as
  \begin{align*}
  \eta(y) &= \int_{\mathcal{X}} k(y,x) \,\mathrm{d}\PM(x) \\
  &= \ip{ \phi(y), \mu[\PM]},
  \end{align*}
 where the kernel embedding is defined as follows.

 \begin{Definition}
 The kernel embedding or kernel mean map $\mu[\PM]$ associated with the continuous, bounded and positive-definite kernel function $k$ is
 \begin{align*}
\mu[\PM] &= \int_{\mathcal{X}} \phi(x) \,\mathrm{d}\PM(x), 
 \end{align*}
 where $\PM$ is a Borel probability measure on $\mathcal{X}$.
 \end{Definition}
 To facilitate the further analysis, we assume that the kernel $k$ is measurable and bounded such that $\mu[\PM]$ exists for all $\PM \in \mathcal{M}_{+}^1(\mathcal{X})$~\autocite{sriperumbudur2011universality}. 
 Since the underlying distribution $\PM$ is in general unknown and only a 
 set of $n\in \Nat$ samples $\SetDef{x_1,\dotsc,x_n}$ from $\PM$ is available for analysis, 
 the empirical measure
 \begin{align*}
 \pmeasureP_n = \frac{1}{n}\sum_{i=1}^n \delta_{x_i},
 \end{align*}
 act as a surrogate, where $\delta_{x}$ is the Dirac measure.
 $\pmeasureP_n$ can be used to construct an approximation $\mu[\PM_n]$ of $\mu[\PM]$ as
 \begin{align*}
 \mu[\PM] \approx  \mu[\PM_n] = \int_{\mathcal{X}} \! \phi(x) \, \dd \pmeasureP_n(x) = \frac{1}{n}\sum_{i=1}^n \phi(x_i)
 \end{align*}
 which is called \emph{empirical kernel embedding}~\autocite{Smola2007}. 
 
 The consequence of the equation above is, that the empirical kernel embedding $\mu[\PM_n]$ has a computational complexity with linear dependence on $n$ and responsible for the \emph{linear} EXPoSE training time. Next, we will look at the EXPoSE classifier from the perspective of a stochastic optimization problem to deliver an $\epsilon$-accurate approximation of $\mu[\PM]$ in \emph{constant} time. A reduction of the computationally complexity from linear to constant for the empirical kernel mean map has significant impact on a variety of applications based on the kernel embedding
 such as for example statistical hypotheses testing~\autocite{Gretton2012} or independence testing~\autocite{gretton2005measuring}.

However the main focus of this work is to improve the EXPoSE training time from linear to constant.
 
 \section{Stochastic Optimization}
 
 This sections derives the stochastic optimization problem together with some general conditions which will be necessary at a later stage. 
 Obviously $\mu[\PM] \in \mathcal{H}$ is the solution of the following unconstrained optimization problem
 \begin{align*}
  \min_{w \in \mathcal{H}} g(w) &= \min_{w \in \mathcal{H}} \norm{ \mu[\PM] - w }_{\mathcal{H}}^2 \\
  &= \min_{w \in \mathcal{H}}  \ip{w,w} -2 \ip{\mu[\PM],w} + \ip{\mu[\PM],\mu[\PM]}  \\
  &= \min_{w \in \mathcal{H}}  \frac{1}{2}\ip{w,w} -\ip{\mu[\PM],w}.
 \end{align*}
 This is equivalent to the \emph{stochastic optimization problem}, where we minimize over the expectation of an objective function
 \begin{align*}
  \min_{w \in \mathcal{H}} \Exp{f(w)} = \min_{w \in \mathcal{H}} \int_{\mathcal{X}} \! f(w) \, \dd \pmeasureP(x),
 \end{align*}
with
 \begin{align*}
	f(w) = \frac{1}{2}\ip{w,w} -\ip{\phi(X),w},
 \end{align*} 
 where the expectation is taken with respect to the random variable $X$.
 
 We will assume that we can generate independent samples from $\PM$ and furthermore require an oracle which returns a \emph{stochastic subgradient} $\tilde{\nabla} f(w)$ of $f$ at $w$. A stochastic subgradient has the property that 
 \begin{align*}
\Exp{\tilde{\nabla} f(w)} = \nabla f(w) \in \partial f(w)
 \end{align*}
which means its expectation is equal to a subgradient $\nabla f(w)$. Here $\partial f(w)$ denotes the set of all subgradients at $w$ called the \emph{subdifferential} which is a subset of the dual $\mathcal{H}^*$ of $\mathcal{H}$ defined by
\begin{align*}
\partial f(x) =  \SetDef[\big]{\xi^* \in \mathcal{H}^* \given f(y) - f(x) \geq \xi^*(y-x)}.
\end{align*}
 
  \begin{Proposition}
  The random variable 
  \begin{align*}
  \tilde{\nabla} f(w) = w - \phi(X)
  \end{align*}
  is a stochastic unbiased gradient of $f$ at $w$.
  \end{Proposition}
  \begin{Proof}
  The expectation of $\tilde{\nabla} f(w)$ is given by
  \begin{align*}
  \Exp{\tilde{\nabla} f(w)} &= \int  w - \phi(X) \dd \PM(x) \\
    &= w - \mu[\PM] \in \partial f(w)
  \end{align*} 
  which is a stochastic unbiased (sub)gradient by definition.
  \end{Proof}

We are going to solve this optimization problem with the \emph{stochastic approximation} algorithm~\autocite{robbins1951stochastic} described next.
 
\subsection{Stochastic Approximation}

Let $\mathcal{H}$ be a Hilbert space, $H \subseteq \mathcal{H}$ be a subset and $f\colon H \to \Real$ some objective function. Furthermore let
\begin{align*}
\Pi_H(w) = \argmin_{v \in H} \norm{w-v}_\mathcal{H}
\end{align*}
be the metric projection operator.
$\Pi_H$ is in general nonexpanding such that
\begin{align*}
\norm{\Pi_H(w) - \Pi_H(w')}_\mathcal{H} \leq \norm{w-w'}_\mathcal{H}
\end{align*}
holds.
Then the classic stochastic approximation algorithm~\autocite{robbins1951stochastic} creates the sequence $(w_t)$ as
\begin{align*}
w_{t+1} = \Pi_H\big( w_t - \gamma_t \tilde{\nabla} f(w_t) \big),
\end{align*}
to solve the stochastic optimization problem
\begin{align*}
  \min_{w \in H} \Exp{f(w)}
\end{align*}
starting at some $w_1 \in H$.
Here $(\gamma_t)$ is a sequence of positive step sizes and the optimal solution to the problem is denoted by $w^\star$.
 
\Citeauthor{nemirovski2009robust} \autocite{nemirovski2009robust} considered $\mathcal{H} = \Real^d$ and showed that stochastic approximation can obtain a $\mathcal{O}(1/t)$ convergence rate if the objective function $f$ is differentiable and $\alpha$-strongly convex on $H$. Here, $\alpha$-strongly convex means there exists a constant $\alpha > 0$ such that
\begin{align*}
f(y) \geq f(x) + \ip{\nabla f(x),y-x} + \frac{1}{2}\alpha\norm{y-x}^2_\mathcal{H}
\end{align*}
for all $x,y \in H$. An additional requirement is that the stochastic subgradient has to be bounded in expectation 
\begin{align*}
 \Exp[\big]{\norm{\tilde{\nabla} f(w)}^2_\mathcal{H}} \leq M^2 \quad \forall w \in H, M > 0
\end{align*}
and the step sizes need to be $\gamma_t = \frac{\theta}{t}$ for some $\theta > \frac{1}{2\alpha}$.
Under these conditions, \citeauthor{nemirovski2009robust} demonstrated that
\begin{align}\label{eq:nemirovski_w}
 \Exp[\big]{\norm{w_t - w^\star}^2_\mathcal{H}} \leq \frac{Q(\theta)}{t}
\end{align}
where
\begin{align*}
Q(\theta) = \max\SetDef*{\theta^2 M^2 (2\alpha\theta-1)^{-1}, \norm{w_1 -w^\star}^2_\mathcal{H}}.
\end{align*}
Furthermore, if the gradient is Lipschitz continuous, i.e. there is a constant $\beta>0$ such that 
\begin{align*}
\norm{\nabla f(x) - \nabla f(y)}_\mathcal{H} \leq \beta \norm{x-y}_\mathcal{H}
\end{align*} 
for all $x,y \in H$, then 
\begin{align}\label{eq:nemirovski_f}
 \Exp[\big]{f(w_t) - f(w^\star)} \leq \frac{1}{2} \frac{\beta Q(\theta)}{t}.
\end{align}
For Lipschitz continuous strongly convex functions, the $\mathcal{O}(1/t)$ rate of convergence is unimprovable \autocite{agarwal2009information}.

We will see that the bound from \citeauthor{nemirovski2009robust} does also hold when $\mathcal{H}$ is a (possibly infinite-dimensional) Hilbert space as in the problem considered in this work. However, some care has to be taken since, unlike in finite-dimensional spaces, being closed and bounded does not imply that a set is compact when $\mathcal{H}$ is infinite-dimensional. We also refer to \autocite{juditsky2010primal} for a discussion on primal-dual subgradient methods in non-Euclidean spaces.

\section{Stochastic Optimization of EXPoSE} 
In this section we show the existence and uniqueness of a solution for the previously defined stochastic optimization problem of EXPoSE and also that it meets all requirements for a $\mathcal{O}(1/t)$ convergence rate.

In the following let $\mathcal{H}$ be a RKHS space with a bounded kernel $k$ such that $\norm{k(x,y)} \leq M^2$. 
Let $H \subseteq \mathcal{H}$ be a weakly sequentially closed and bounded set with $\norm{H}_\mathcal{H} \leq M$.
It is not hard to show the existence of a minimizer of 
\begin{align}
  \min_{w \in H} \Exp{f(w)} = \min_{w \in H} \int \! \frac{1}{2}\ip{w,w} -\ip{\phi(X),w} \, \dd \pmeasureP,\label{eq:objective}
\end{align}
since we already know the solution 
$w^\star = \mu[\PM]$
assuming that $\mu[\PM] \in H$. This assumption holds since
\begin{align*}
\norm{\mu[\PM]}_{\mathcal{H}}^2 &= \norm[\Big]{\int_{\mathcal{X}} \phi(x) \,\mathrm{d}\PM(x) }_{\mathcal{H}}^2 \\
& \leq  \int_{\mathcal{X}} \norm{\phi(x)}_{\mathcal{H}}^2 \,\mathrm{d}\PM(x)\\
& = \int_{\mathcal{X}} k(x,x) \,\mathrm{d}\PM(x)\\ 
& \leq M^2.
\end{align*}
 This solution is also unique. The proof requires $f(w)$ to be strongly convex, which is subject of the following property:
\begin{Proposition}\label{thm:strongly_convex}
   The objective function $f(w)$ is $\alpha$-strongly convex and its gradient is $\beta$-Lipschitz with $\alpha = \beta = 1$.
\end{Proposition}
\begin{Proof}
A function $f$ is $\alpha$-strongly convex if and only if $w\mapsto f(w)-\frac{\alpha}{2}\norm{w}^2_\mathcal{H}$ is convex.
\begin{align*}
f(w) - \frac{1}{2}\norm{w}^2_\mathcal{H} &= \frac{1}{2}\ip{w,w} -\ip{\phi(X),w} - \frac{1}{2}\norm{w}^2_\mathcal{H}\\
&= -\ip{\phi(X),w}
\end{align*}
which is convex in $w$. Hence $\alpha = 1$.

Furthermore 
$Df(w)\colon z \mapsto \ip{w-\phi(X),z}$ is the Fréchet derivative of $f$ at $w$ since
\begin{align*}
\lim_{h \to 0} \frac{\norm{f(w+h) - f(w) - \ip{Df(w)\vert h}}}{\norm{h}_\mathcal{H}} = 0
\end{align*}
with dual pairing $\ip{\cdot \vert \cdot}$. The gradient $\nabla f(w) = w-\phi(X)$ is $\beta$-Lipschitz since
\begin{align*}
\norm{\nabla f(w) - \nabla f(v)}_\mathcal{H} &= \norm{w - \phi(X)- v + \phi(X)}_\mathcal{H}\\
&= \norm{w-v}_\mathcal{H}
\end{align*} 
for all $w,v \in H$ due to Riesz representation.
\end{Proof}

Besides the existence of a minimizer, its uniqueness plays an important role. The sufficient conditions for $w^\star$ to be unique are given by \autocite[Corollary 2.19]{peypouquet2015convex} which states the following:
\begin{Corollary}
Let $\mathcal{H}$ be reflexive. If $f\colon\mathcal{H}\to\Real \cup \SetDef{+\infty}$ is proper, convex, coercive and lower-semicontinuous, then $\argmin f$ is nonempty and weakly compact. If, moreover $f$ is strictly convex, then $\argmin f$ is a singleton.
\end{Corollary} 
\begin{Proof}[Proof: $w^\star$ is unique]
All Hilbert spaces are reflexive. Since $f$ is continuous, proper ($\dom (f) \neq \SetDef{}$) and strongly convex it is also convex, coercive and lower-semicontinuous.
\end{Proof} 
 
Next we state the two main theorems of this paper.
\begin{Theorem}\label{thm:exp_w}
Using the sequence
\begin{align*}
w_{t+1} = \Pi_H\big( w_t - \gamma_t \tilde{\nabla} f(w_t) \big),
\end{align*}
with $f$ given by \cref{eq:objective} we have 
\begin{align*}
 \Exp[\big]{\norm{w_t - w^\star}^2_\mathcal{H}} \leq \frac{M^2}{t}
\end{align*}
for all $t \in \Nat$.
\end{Theorem}
\begin{Proof}
Since $Q(\theta)$ attains its optimal value at $\theta = 1/\alpha$ we get from \cref{eq:nemirovski_w} that
\begin{align*}
 \Exp[\big]{\norm{w_t - w^\star}^2_\mathcal{H}} \leq t^{-1} \max\SetDef*{\alpha^{-2} M^2 , \norm{w_1 -w^\star}^2_\mathcal{H}} 
\end{align*}
and we have
\begin{align*}
\Exp*{\norm{w_1 - w^\star}_\mathcal{H}^2} \leq \frac{M^2}{\alpha^2}
\end{align*}
since strong convexity implies
\begin{align*}
\ip{w-w^\star,\nabla f(w)} &\geq \alpha \norm{w-w^\star}_\mathcal{H}^2 \\
\ip{w-w^\star,\nabla f(w)}^2 &\geq \alpha^2 \norm{w-w^\star}_\mathcal{H}^4
\end{align*}
and by Cauchy-Schwartz inequality we get
\begin{align*}
\norm{w-w^\star}_\mathcal{H}^2 \cdot \norm{\nabla f(w)}_\mathcal{H}^2 &\geq  \ip{w-w^\star,\nabla f(w)}^2    
\end{align*}
which yields
\begin{align*}
\norm{w-w^\star}_\mathcal{H}^2 \cdot \norm{\nabla f(w)}_\mathcal{H}^2 &\geq \alpha^2 \norm{w-w^\star}_\mathcal{H}^4 \\
\norm{\nabla f(w)}_\mathcal{H}^2 &\geq \alpha^2 \norm{w-w^\star}_\mathcal{H}^2
\end{align*}
for all $w$. Taking exponents on both sides and the bound $\norm{\nabla f(w)}_\mathcal{H}^2 \leq M^2$ 
we get 
\begin{align*}
 \Exp[\big]{\norm{w - w^\star}^2_\mathcal{H}} \leq \alpha^{-2} M^2
\end{align*}
which concludes the proof using $\alpha = 1$ (\cref{thm:strongly_convex}).
\end{Proof} 
 
Notice that $\norm{\nabla f(w)}_\mathcal{H}^2 \leq M^2$ does indeed hold since $\nabla f(w) = w - \phi(X) \in H$. The following theorems describes the convergence rate of the objective function $f$ in terms of the number of iterations $t$.
 
\begin{Theorem}
Under the prerequisites of \cref{thm:exp_w} it holds that
\begin{align*}
\Exp[\big]{f(w_t) - f(w^\star)} \leq \frac{1}{2} \frac{M^2}{t}.
\end{align*}
\end{Theorem}
\begin{Proof}
Using \cref{eq:nemirovski_f} and the bound for $Q(\theta)$ derived before yields the desired result.
\end{Proof}

We showed above that (in expectation) the distance between the optimal objective $f(w^\star)$ and $f(w_t)$ decays as $\mathcal{O}(1/t)$.
Another question is how this effects the EXPoSE decision rule $\eta(y) = \ip{ \phi(y), \mu[\PM]}$. By definition and the application of the Cauchy–Schwarz inequality it holds that
\begin{align*}
   \norm{  \ip{ \phi(y), \mu[\PM]} - \ip{ \phi(y), w_t} } &=  \norm{  \ip{ \phi(y), \mu[\PM] - w_t}} \\
   & \leq \norm{\phi(y)}_{\mathcal{H}} \cdot \norm{\mu[\PM] - w_t}_{\mathcal{H}}
\end{align*}
for all $y \in \mathcal{H}$. Taking expectations yields
\begin{align*}
   \Exp[\big]{\norm{  \ip{ \phi(y), \mu[\PM]} - \ip{ \phi(y), w_t} }^2} 
   & \leq \norm{\phi(y)}_{\mathcal{H}}^2 \frac{M^2}{t}
\end{align*}
for all $t \in \Nat$. 

\begin{algorithm}
\caption{EXPoSE using Stochastic Optimization
    \label{alg:expose_so}}
\begin{algorithmic}[1]
\Require 
\State $T:$ the number of iterations \textit{or}  $\epsilon:$ accuracy
\Main
   \State Set $w_1 \gets 0$
   \For{$t \gets 1, 2, \dotsc, T$}
   \State Sample $x_t$ uniformly from $\PM$
   \State Set $\gamma_t \gets \frac{1}{t}$   
   \State Set $\tilde{\nabla} f(w_t) \gets w_t-\phi(x_t)$
   \State Update $w_{t+1} \gets w_t - \gamma_t \tilde{\nabla} f(w_t)$
   \State Project $w_{t+1} \gets  
   w_{t+1} \cdot \max\SetDef{1,M\norm{w_{t+1}}}^{-1}$
   \EndFor
   \State \Return{$w_{T+1}$}
\end{algorithmic}
\end{algorithm}

The stochastic optimization procedure for EXPoSE is summarized in \cref{alg:expose_so}. Please note that the stochastic optimization procedure presented here is relatively simple and requires only a few lines of code to implement. It also does not introduce additional parameters since the optimal step-size in known. Step-sizes are crucial and difficult to determine in most optimization algorithms as they have a significant effect on the results. The bound $M$ of the kernel is typically known and the number of iterations $T$ determines the computing time and accuracy. Alternatively, the number of iterations $T$ can be calculated given a desired accuracy $\epsilon$ using \cref{thm:exp_w}. The projection operator $\Pi_H(w)$ in the last step takes a form which can efficiently be computed, projecting $w$ onto the sphere $H$.

We emphasize that the stochastic optimization procedure introduce here does not improve on the $\mathcal{O}(1/\sqrt{t}))$ convergence rate of the empirical kernel mean map as demonstrated in \cref{thm:exp_w}, but introduces a methodology to reduce the computational complexity from linear to constant.

\subsection{Convergence of EXPoSE}
Since $w \rightarrow w^\star$ converges, this implies also the weak convergence \autocite{peypouquet2015convex} from $w \rightharpoonup w^\star$ namely
\begin{align*}
\lim_{t\to \infty} \ip{u,w_t} = \ip{u,w^\star}, \quad \forall u \in H
\end{align*}
and especially
\begin{align*}
\lim_{t\to \infty} \ip{\phi(y),w_t} = \ip{\phi(y),\mu[\PM]}, \quad \forall y \in \mathcal{X}
\end{align*}
which justifies the use of $w_t$ as a surrogate for $\mu[\PM]$.

\subsection{Regularization}

We would like to mention that the
reformulation of EXPoSE as an optimization problem also introduces the opportunity to add constraints or similar properties to the objective function.
One approach is to define a general \emph{regularizer} $\lambda\Omega(w)$ on $H$ replacing $\frac{1}{2}\ip{w,w}$ in \cref{eq:objective} which yields
\begin{align*}
  \min_{w \in H} \Exp{f(w)} = \min_{w \in H} \int \! \lambda\Omega(w) -\ip{\phi(X),w} \, \dd \pmeasureP
\end{align*}
with some regularization parameter $\lambda \geq 0$.
An example would be to add a roughness penalty to the space of functions setting
\begin{align*}
   \lambda\Omega(w) = \lambda\ip{D^2w,D^2w}
\end{align*}
where $D$ denote the differential operator. 
Another possibility is to places a sparsity constraint on $w$. If $\mathcal{H}$ admits it, we can use
\begin{align*}
	\lambda\Omega(w) = \lambda\norm{w}_1,
\end{align*}
where $\norm{\cdot}_1$ is the $l_1$-norm. 

The disadvantage of other objective functions is, that these are in general not strongly-convex and hence yielding a slower convergence rate and may require additional parameters which are difficult to tune.

\section{Experimental Evaluation}

We present experimental results demonstrating the benefit of the proposed approach. Since the true distribution $\PM$ is often unknown and a closed form solution of $\mu[\PM]$ is not available, we will use the empirical distribution $\pmeasureP_n$ as its surrogate in the objective function and measure the behavior of 
\begin{align*}
\norm{w_t-\mu[\pmeasureP_n]}_{\mathcal{H}}
\end{align*} 
as $t$ increases. For sufficiently large sample sizes $n$ we can expect $\mu[\pmeasureP_n]$ to be a good proxy for $\mu[\PM]$ by the law of large numbers.
Besides the convergence of the model $w_t \to \mu[\PM]$, we will examine and compare the anomaly detection scores 
\begin{align*}
\eta_n(y) &= \ip{\phi(y), \mu[\pmeasureP_n]} \quad \text{and} \\
\eta_t(y) &= \ip{\phi(y), w_t}
\end{align*} 
calculated by the empirical distribution (which is the original EXPoSE predictor proposed in \autocite{Schneider2015a}) and the stochastic optimization approximation, respectively.

\subsection{Approximate Feature Maps}

While it is theoretically possible to calculate quantities like $\norm{w_t-\mu[\pmeasureP_n]}_{\mathcal{H}}$ for any kernel $k$, this is extremely slow and intractable for most large-scale datasets. For datasets with a small sample size $n$ we cannot expect $\mu[\pmeasureP_n]$ to be a good proxy for $\mu[\PM]$. We therefore omit an experiment with explicit features as we either cannot compute $\mu[\pmeasureP_n]$ (large $n$) or $\mu[\pmeasureP_n]$ is not a good estimate for $\mu[\PM]$ (small $n$).

In order to overcome this problem, EXPoSE exploits the idea of \emph{approximate feature maps} for its computational efficiency.
The aim is to find approximations $\hat{\phi}\colon\mathcal{X}\to\Real^r$ of $\phi$ such that
\begin{align*}
k(x,y) \approx \ip{\hat{\phi}(x), \hat{\phi}(y)}
\end{align*}
for all $x,y \in \mathcal{X}$ and $r\in\Nat$. We will utilize the Random Kitchen Sinks (RKS) approach \cite{Rahimi2007,Rahimi2008} which is based on Bochner's theorem for translation invariant kernels (such as the Gaussian RBF, Laplace, Mat\'ern covariance, etc.). For example in the following experiments we will use the Gaussian RBF kernel 
$k(x,y) = \exp\big(-\frac{1}{2\sigma^2}\norm{x-y}^2\big)$,
which can be approximated by
\begin{align*}
Z &\in \mathbb{R}^{r \times d} \text{ with } Z_{ij} \sim \mathcal{N}(0,\sigma^{2}) \\
\hat{\phi}(x) &= \frac{1}{\sqrt{r}}\exp(\im Zx), 
\end{align*}
where $d$ is the dimension of $\mathcal{X} \subseteq \Real^d$. The parameter $r \in \Nat$ determines the number of kernel expansions and is typically around \num{20000}. The specific choice of approximate feature map \emph{does not} affect the previous theoretical analysis and other feature map approximations \autocite{li2010random,vedaldi2012efficient,kar2012random} can be used as well.

\subsection{Datasets}

The following datasets, which all have purposely very different feature characteristics, are used to perform anomaly detection. We refer to \autocite{Schneider2015a} for a detailed description of the datasets and feature characteristic.
\begin{itemize}
\item The MNIST database contains \num{70000} images of handwritten digits. Using the raw pixel values yield an input space dimension of \num{784}.
\item  KDD-CUP 99 is an intrusion detection dataset which contains \num{4898431} connection records of network traffic. As in \autocite{Schneider2015a} we rescale the $34$ continuous features to $[0,1]$ and apply a binary encoding for the $7$ symbolic features.
\item The third dataset contains \num{600000} instances of the \emph{Google Street View House Numbers} (SVHN) \cite{Netzer2011} where we use the \emph{Histogram of Oriented Gradients} (HOG) with a cell size of $3$ to get a $2592$-dimensional feature vector.
\end{itemize}
The kernel bandwidth $\sigma^2$ used for these datasets are \num{7.0}, \num{5.6} and \num{7.8} respectively, which we found to yield a reasonable anomaly detection performance.

Since SVHN and MNIST are multi-class and not anomaly detection datasets 
we use the digit $1$ as \emph{normal} class and all other digits as \emph{anomaly} instances.\footnote{A different normal/anomaly setup had no significant impact on the experimental results.}
At each iteration of \cref{alg:expose_so} we uniformly choose an instance from the (training) dataset not used previously. We then update the model $w_t$ according to the algorithm. Every $200$ iterations, $w_t$ is used to calculate an anomaly detection score for \num{10000} dedicated random instances of the (test) dataset using the full model $\eta_n(y)$ and the stochastic optimization approximation $\eta_t(y)$.

\subsection{Discussion}

\begin{figure*}
\begin{subfigure}[b]{.31\linewidth}
\caption{MNIST}
\includegraphics{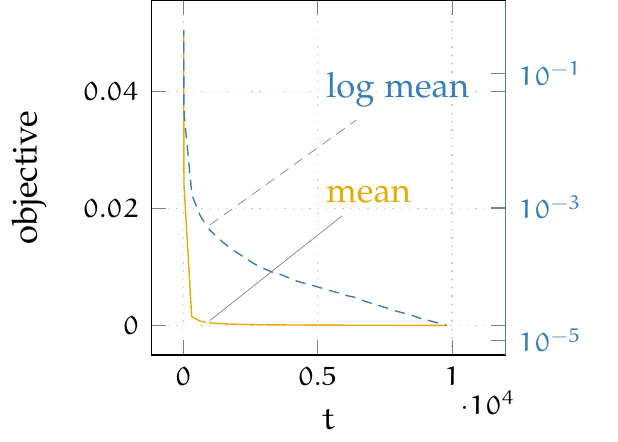}
\end{subfigure}%
\begin{subfigure}[b]{.31\linewidth}
\caption{SVHN}
\includegraphics{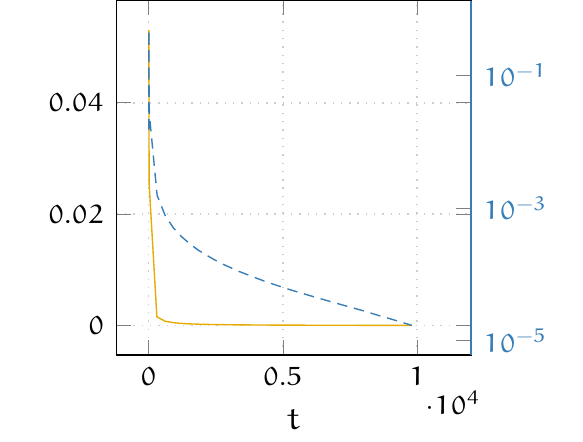}
\end{subfigure}%
\begin{subfigure}[b]{.31\linewidth}
\caption{KDD CUP 99}
\includegraphics{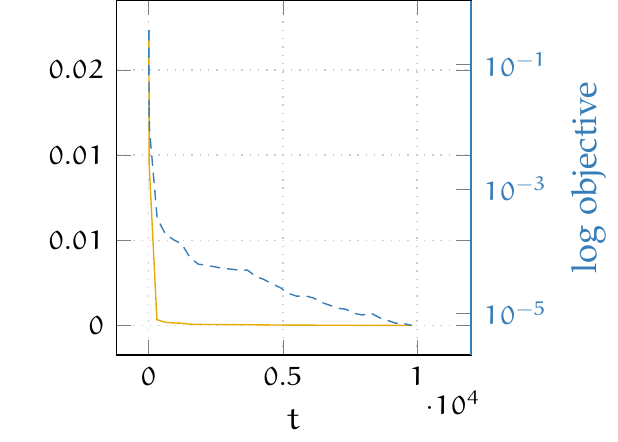}
\end{subfigure}%

\begin{subfigure}[b]{.31\linewidth}
\includegraphics{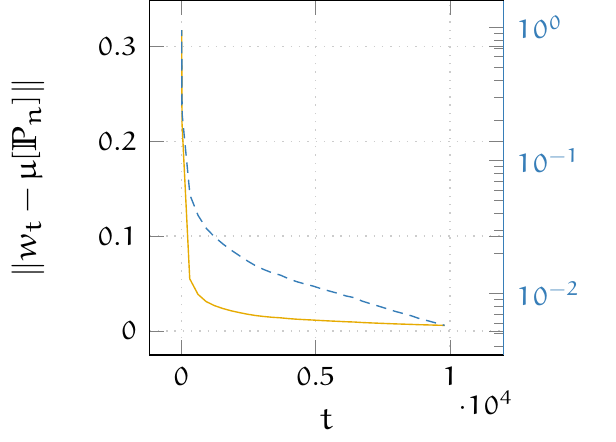}
\end{subfigure}%
\begin{subfigure}[b]{.31\linewidth}
\includegraphics{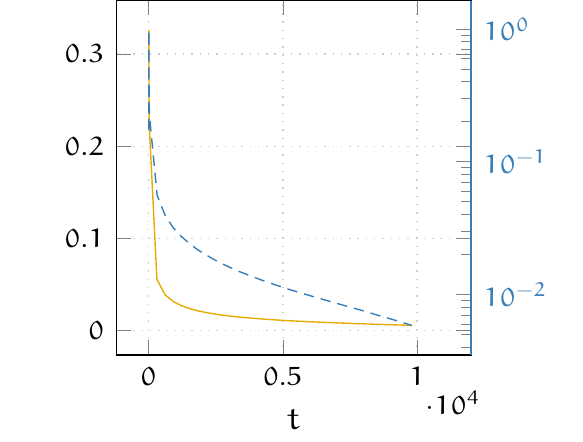}
\end{subfigure}%
\begin{subfigure}[b]{.31\linewidth}
\includegraphics{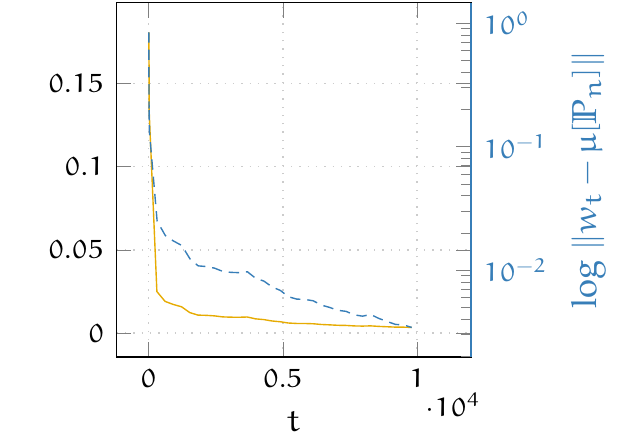}
\end{subfigure}%

\begin{subfigure}[b]{.31\linewidth}
\includegraphics{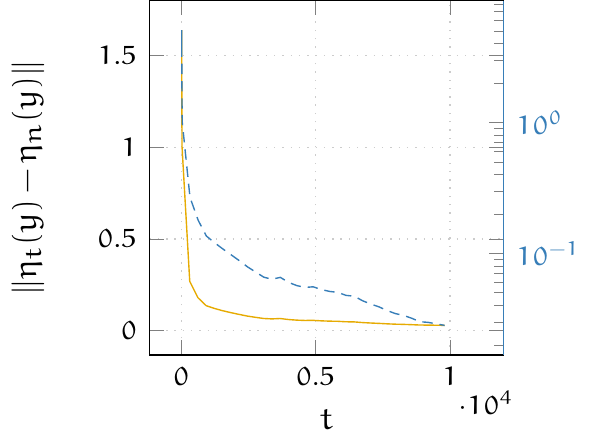}
\end{subfigure}%
\begin{subfigure}[b]{.31\linewidth}
\includegraphics{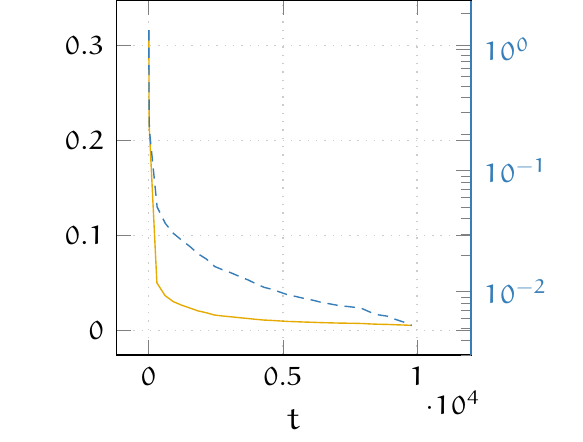}
\end{subfigure}%
\begin{subfigure}[b]{.31\linewidth}
\includegraphics{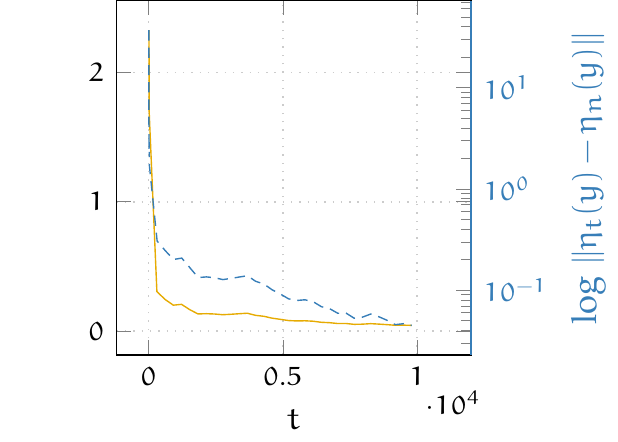}
\end{subfigure}%

\begin{subfigure}[b]{.31\linewidth}
\includegraphics{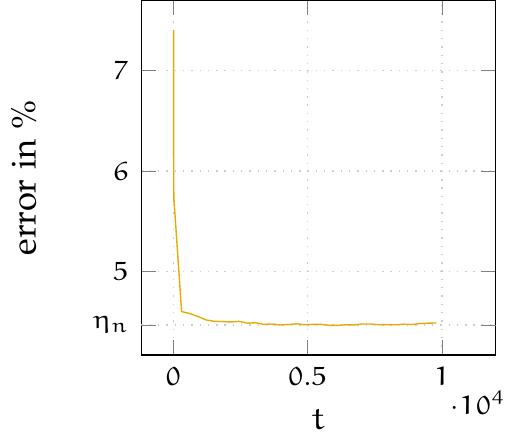}
\end{subfigure}%
\begin{subfigure}[b]{.31\linewidth}
\includegraphics{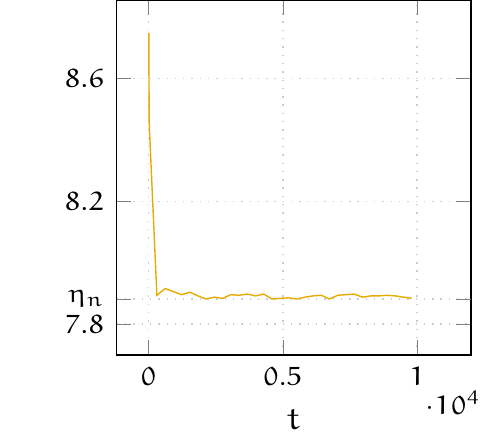}
\end{subfigure}%
\begin{subfigure}[b]{.31\linewidth}
\includegraphics{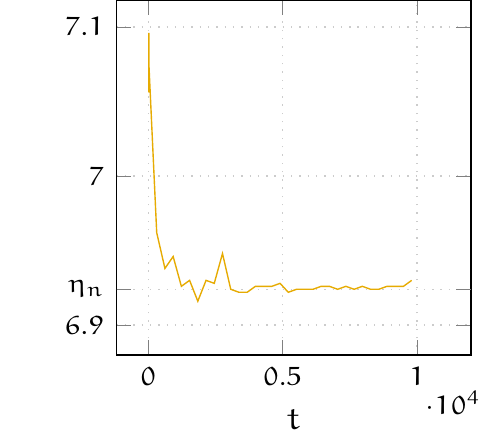}
\end{subfigure}%
\caption{Evaluation of the stochastic optimization approach for EXPoSE. The datasets are organized in columns. The first row illustrates the difference between objective functions $f(w_t) - f(w^\star)$. In the second row we show the deviation of $w_t$ from $\mu[\pmeasureP_n]$ as $\norm{w_t-\mu[\pmeasureP_n]}$. In the third row we plotted the difference in scores $\norm{\eta_t(y)- \eta_n(y)}$, averaged over all query points $y$ in the test dataset. The last row shows the anomaly detection performance of EXPoSE.
In all figures, the solid line is the mean over $10$ experiments and on the second $y$-axis on the right we show the same curve (dashed) on a logarithmic scale when appropriate.
}
\label{fig:SO_experimental_results}
\end{figure*}

The experimental results with approximate feature maps are shown in \cref{fig:SO_experimental_results}. The first row contains traces of the objective function $f(w_t) - f(w^\star)$, where $w^\star \approx \mu[\pmeasureP_n]$ for all three datasets. The stochastic optimization algorithm already reaches a reasonable low objective after a few hundred iterations. A further improvement is only visible on a logarithmic scale (dashed blue) on the second $y$-axis on the right. 
More important, we observe a similar effect in the second row when comparing $\norm{w_t-w^\star}$. We get near to $w^\star$ relatively fast, but it takes much more samples to estimate $w^\star$ with a high accuracy. However, we will see that a high accuracy estimation is necessary for a good anomaly detection performance.
To measure the anomaly detection rate, we first plug $w_t$ and $w^\star$ into the EXPoSE estimators $\eta_t(y)$ and $\eta_n(y)$ respectively and calculate scores for all instances in the test dataset. The difference of these scores are shown in row number three. We see again, that the stochastic optimization approximation $\eta_t(y)$ yields similar scores as the full $\eta_n(y)$.
The last row illustrates the development of the classification error as more iterations are performed\footnote{The prediction score threshold is determined by means of cross-validation.}. \emph{After only a few hundred iterations $\eta_t(y)$ reaches the same classification error as the original EXPoSE predictor $\eta_n(y)$.}
This confirms that a high accuracy approximation of $w^\star$ does not necessarily lead to a better predictor. The key is, that for a given $\epsilon$ we can reach $\norm{w_t-w^\star} <\epsilon$ in a fixed number of iterations, \emph{independent} of the dataset size $n$ which reduced the computational complexity from $\mathcal{O}(n)$ to $\mathcal{O}(1)$.

We emphasize that, unlike other regularized risk minimization problems, EXPoSE does not have a regularization parameter. This is important as the authors of Pegasos noticed that \textcquote{Shalev-Shwartz2007}{\textelp{} the runtime to achieve a predetermined suboptimality threshold would increase in proportion to $\lambda$ \textins{the regularization parameter}. Very small values of $\lambda$ (small amounts of regularization) result in rather long runtimes}.

\section{Conclusion}

In this work we cast the EXPoSE anomaly detection algorithm into a stochastic optimization problem. This enables us to fine an $\epsilon$-accurate approximation of the kernel mean map $\mu[\PM]$ in \emph{constant time}, independent of the training dataset size $n$. In particular, this approximations reduces the computational complexity of EXPoSE and the empirical kernel mean map from the previous $\mathcal{O}(n)$ to $\mathcal{O}(1)$ whenever an $\epsilon$-accurate estimation is sufficient.
More precisely, we are able to determine the number of necessary stochastic optimization iterations $T$ for a user defined error threshold $\epsilon$ such that $\norm{w_T-w^\star} <\epsilon$. The intuition is that a very high accuracy estimation $w^\star$ does not necessarily result in a better anomaly detection performance and hence there is no benefit in spending more computational resources. This intuition is also confirmed experimentally on three large-scale datasets, where we reach the same anomaly detection performance long before all data is incorporated into the model.
This is the first time that an optimization routine is used for EXPoSE and we provide a detailed theoretical analysis of this algorithm. 
We emphasize that the proposed algorithm does not introduce any additional parameters which have to be tuned and the gradient descent step-sizes are determined automatically. 
This has significant implications for large-scale applications such as anomaly detection problems and other techniques which are based on the kernel mean embedding.

\printbibliography
\end{document}